\documentclass[10pt,twocolumn,letterpaper]{article}

\usepackage{cvpr}
\usepackage{times}
\usepackage{epsfig}
\usepackage{graphicx}
\usepackage{amsmath}
\usepackage{mathtools}
\DeclarePairedDelimiter\ceil{\lceil}{\rceil}
\DeclarePairedDelimiter\floor{\lfloor}{\rfloor}

\usepackage{amssymb}
\usepackage{xcolor}

\usepackage[breaklinks=true,bookmarks=false]{hyperref}

\cvprfinalcopy 

\def\cvprPaperID{****} 

\begin{document}

\title{What are the Receptive, Effective Receptive, and Projective Fields \\ of Neurons in Convolutional Neural Networks?}

\author{Hung Le\\
University of Central Florida\\
{\tt\small hungle@knights.ucf.edu}
\and
Ali Borji\\
Center for Research in Computer Vision \\
University of Central Florida\\
{\tt\small aborji@crcv.ucf.edu}
}

\maketitle

\begin{abstract}
In this work, we explain in detail how receptive fields, effective receptive fields, and projective fields of neurons in different layers, convolution or pooling, of a Convolutional Neural Network (CNN) are calculated. While our focus here is on CNNs, the same operations, but in the reverse order, can be used to calculate these quantities for deconvolutional neural networks. These are important concepts, not only for better understanding and analyzing convolutional and deconvolutional networks, but also for optimizing their performance in real-world applications. 
\end{abstract}


\section{Definitions}
\label{sec1}
\textbf{Receptive Field (RF):} is a local region (including its depth) on the output volume of the previous layer that a neuron is connected to. This term has been prevalent in Neurosciences since the study of Hubel and Wiesel~\cite{hubel1962receptive} in which they suggested local features are detected in early visual layers of the visual cortex and are then progressively combined to create more complex patterns in a hierarchical manner.

As an example, assume that the input RGB image to a CNN 
has size $[32\times32\times3]$. For a filter size of  $5\times5$, then each neuron in the first convolutional layer will be connected to a $[5\times5\times3]$ region in the input volume. Thus, a total of $5\times5\times3 = 75$ weights (+1 bias parameter) needs to be learned. Notice that RF is a 3D tensor with its depth being equal to the depth of the volume in the previous layer. Here, for simplicity, we discard the depth in our calculation.

\textbf{Effective Receptive Field (ERF):} is the area of the original image that can possibly influence the activation of a neuron. One important point to notice here is that RF and ERF are the same for the first convolutional layer. However, they differ as we moves along the CNN hierarchy. The RF is simply equal to filter size over the previous layer but ERF traces the hierarchy back to the input image and indicates the extent of the input image which can modulate the activity of a neuron. Here, we focus on ERF calculation. It is worth noting that ERF and RF are sometimes used interchangeably (and hence confused) in the computer vision community. 

\textbf{Projective Field (PF):} is the set of neurons to which a neuron projects its output~\cite{lehky1988network}. 

\begin{figure}[t]
\begin{center}
\includegraphics[width=0.7\linewidth]{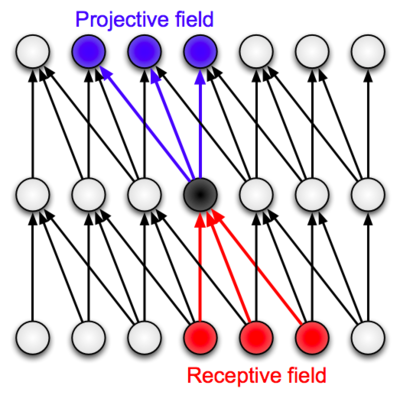}
\end{center}
   \caption{Schematic plot demonstrating receptive and projective fields of a neuron (Borrowed from http://www.scholarpedia.org/article/Projective\_field).}
\label{fig:RFPF}
\end{figure}

Figure~\ref{fig:RFPF} illustrated these definitions.

\section{Calculating the ERF}
In convolutional neural networks~\cite{lecun1998gradient}, the ERF of a neuron indicates which area of the input image is considered by the filter. Calculating the size of the ERF would help choosing suitable filter sizes using domain knowledge for enhancing the performance of CNNs.
\\
There are two ways to calculate the ERF size: 1) Bottom-Up, and 2) Top-Down. Both ways produce the same result. The intermediate values of each approach, however, have different meanings in each case.

\subsection{Bottom-Up Approach}

\begin{figure}[t]
\begin{center}
\includegraphics[width=0.6\linewidth]{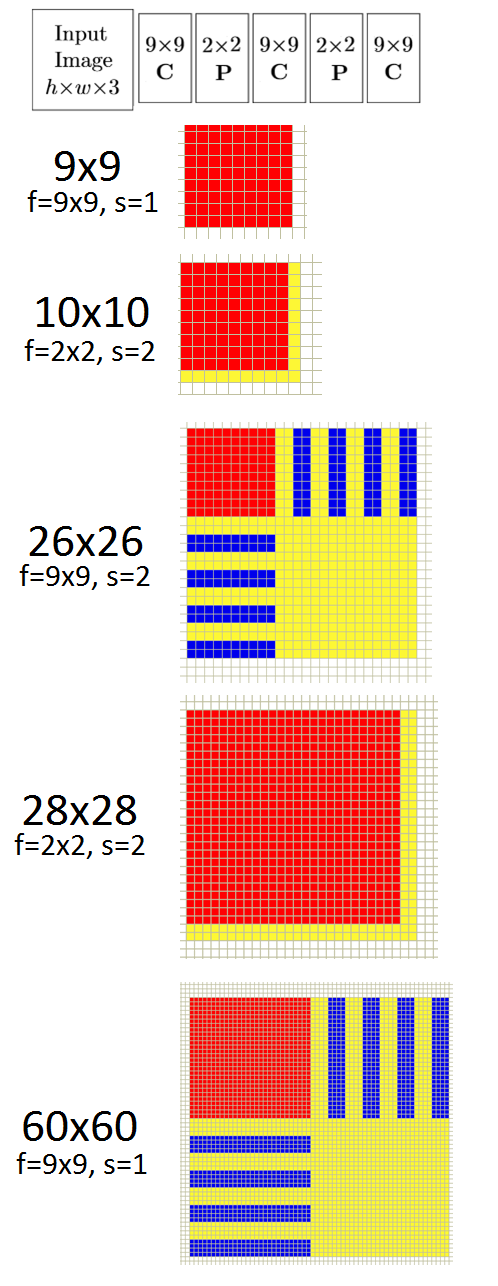}
\end{center}
   \caption{Example of Bottom-up approach for ERF calculation for the network shown in the top row. Red area is the ERF of the lower layer. Yellow and blue are non-overlapped areas, used to indicate how stride affects the calculation of the additional area. In this example, after the first pooling, each additional filter adds 2 pixels to the ERF. After the second pooling, each additional filter adds 4 pixels to the ERF.}
\label{fig:BottomUp}
\end{figure}

The bottom-up approach is the method to calculate the ERF of a neuron at layer k projected on the input image. Let \(R_{k}\) be the ERF of a neuron at layer k. Given the ERF of the previous layer \(R_{k-1}\), where \(R_{0} = 1\) is the ERF at input image layer, the ERF for a neuron at current layer \(R_{k}\) can be computed by adding the non-overlapped-area A to \(R_{k-1}\):
\begin{equation}
R_{k} = R_{k-1} + A
\label{eq:eq1}
\end{equation}
Let \(f_{k}\) represent the filter size of layer k. There are \((f_{k} - 1)\) filters overlap with each other. Since a filter can be convolved with a stride greater than one, it can significantly increase the non-overlapped area. Thus, it is necessary to account for the number of pixels each extra filter contributes to the ERF. Since the stride of the lower layer also affects the ERF of the higher layer, the pixel contributions of all layers must be accumulated. Therefore the non-overlapped area is calculated as:
\begin{equation}
A = (f_{k} - 1)\prod^{k-1}_{i=1}s_i
\label{eq:eq2}
\end{equation}
where \(s_{i}\) is the stride of the layer i. Combining equation \ref{eq:eq1} and equation \ref{eq:eq2}, the ERF can be computed as:
\begin{equation}
R_{k} = R_{k-1} + (f_{k} - 1)\prod^{k-1}_{i=1}s_i
\label{eq:eq3}
\end{equation}

\begin{figure}[t]
\begin{center}
\includegraphics[width=.9\linewidth]{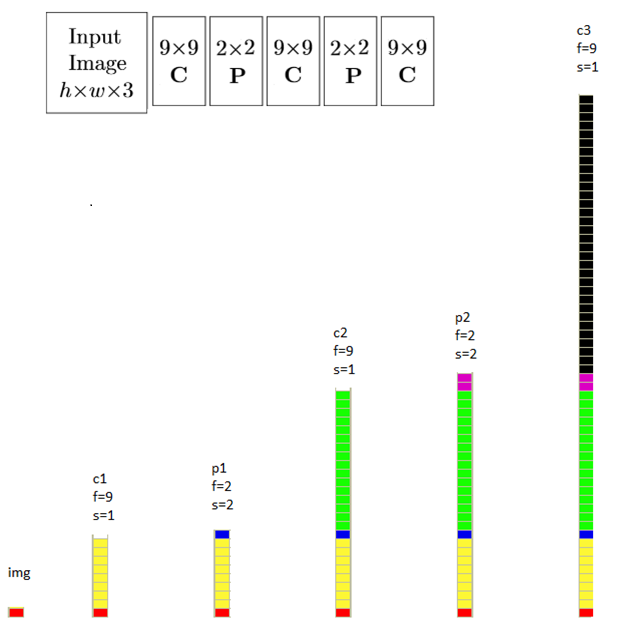}
\end{center}
   \caption{1 dimensional example illustrating how each layer expands the ERF.}
\label{fig:BottomUp1D}
\end{figure}

Figures \ref{fig:BottomUp} and \ref{fig:BottomUp1D} illustrate ERF calculation for a sample architecture. The advantage of bottom-up approach is that it produces the ERF for all layers in one feed-forward pass. 

\subsection{Top-Down Approach}
The computation of ERF in this approach is done via calculating the RF a neuron at layer k projected on the lower layer j, where the RF of the last layer would be the ERF. Given the RF of a neuron at higher layer \(R_{k,j+1}\), if there is no overlap (i.e., stride equal filter size) then the RF of the current layer is:
\begin{equation}
R_{k,j} = R_{k,j+1} f_{j+1}
\label{eq:eq4}
\end{equation}
where \(f_{j+1}\) is the filter size of higher layer. The RF is 1 when \(j = k\). When the filters are overlapped with each other, the overlapped area must be subtracted from the value. Imagine placing down a filter, then every filter being placed after it would have area overlap with the previously placed filter. Since the RF of the higher layer is the one being projected down, the number of filters that overlap with each other would simply be the RF of higher layer minus one:
\begin{equation}
O = R_{k,j+1}-1
\label{eq:eq5}
\end{equation}
The subsequent filter being placed down would be shifted by the stride, thus the overlapped area is dependent on the size of the filter and the stride. Larger strides would yield to less overlap. Larger filters would result in more overlap. The overlapped area of each filter would be the difference of filter size and stride of the higher layer \(s_{j+1}\):
\begin{equation}
A = f_{j+1}-s_{j+1}
\label{eq:eq6}
\end{equation}
Having the number of overlapped filters and the area that each filter overlapped, the RF at the current layer can be computed by combining the equations \ref{eq:eq4}, \ref{eq:eq5}, and \ref{eq:eq6}:
\begin{equation}
R_{k,j} = R_{k,j+1} f_{j+1} - (R_{k,j+1}-1)(f_{j+1}-s_{j+1})
\label{eq:eq7}
\end{equation}
Expanding and simplifying the above equation gives the final top-down equation:
\begin{equation}
R_{k,j} = (R_{k,j+1}-1) s_{j+1} + f_{j+1}
\label{eq:eq8}
\end{equation}

\begin{figure}[t]
\begin{center}
\includegraphics[width=0.6\linewidth]{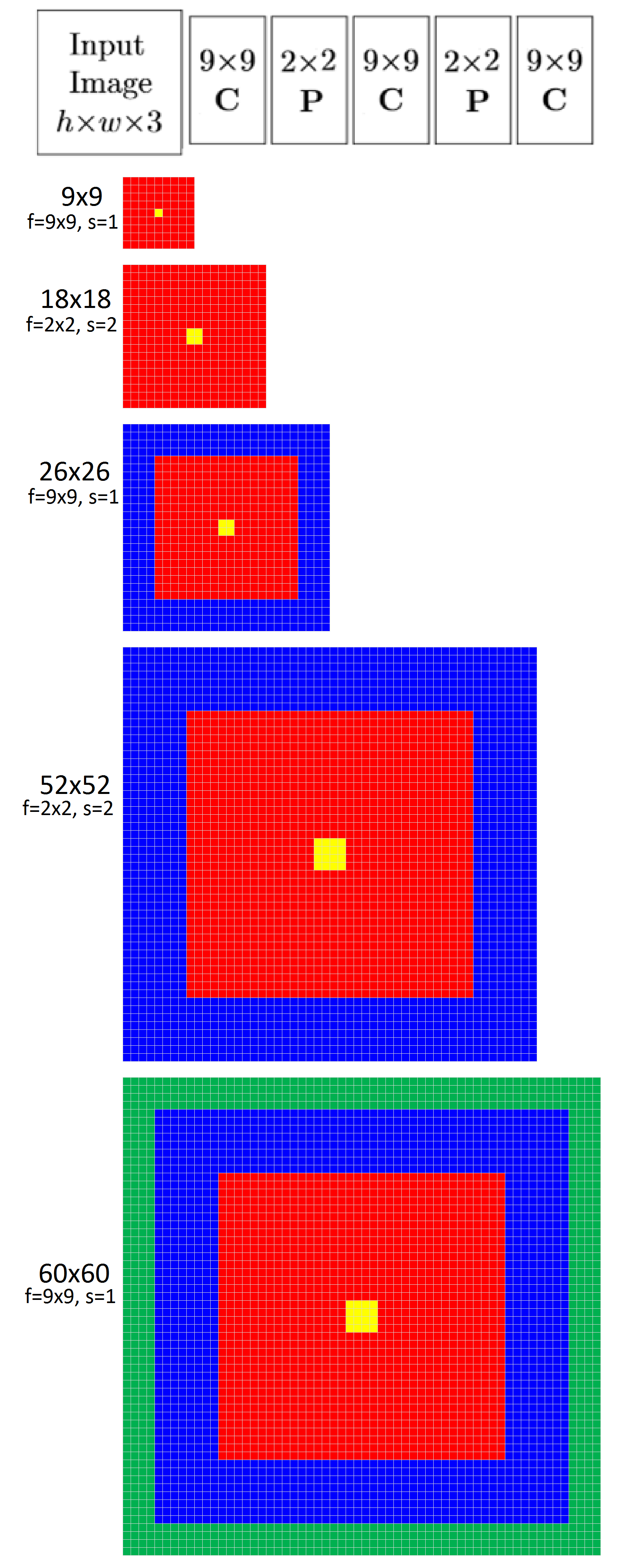}
\end{center}
   \caption{Example of Top-Down approach showing the RF of the last layer ($9\times9$ Convolution) being projected back to the input image. With stride of 2 and filter size of 2 $\times$ 2, the RF is simply doubled in size.}
\label{fig:TopDown}
\end{figure}

\begin{figure}[t]
\begin{center}
\includegraphics[width=.9\linewidth]{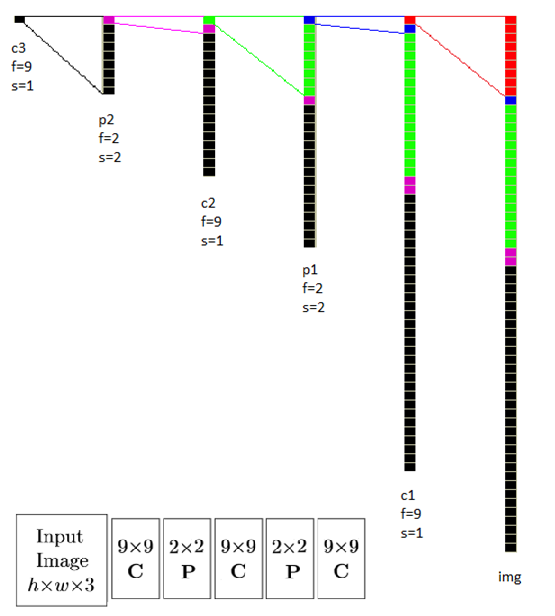}
\end{center}
   \caption{1 dimensional example illustrating how the top-down approach expands the RF through each lower layer.}
\label{fig:TopDown1D}
\end{figure}

The top-down approach is helpful during the analysis as it can be computed relatively quickly. Also, given a point on a filter, it is possible to speculate the nodes that contributed to its output. For the deconvolutional networks, the top-down approach can be used to control the resolution of the output image. Thus instead of using upside down CNN layers, the deconvolutional layers can be designed to incorporate any domain knowledge about the problem. Figures \ref{fig:TopDown} and \ref{fig:TopDown1D} show examples of the progression of RF being projected back to lower layers.

\subsection{Case Study}
\begin{figure*}[t]
\centerline{\includegraphics[width=\textwidth]{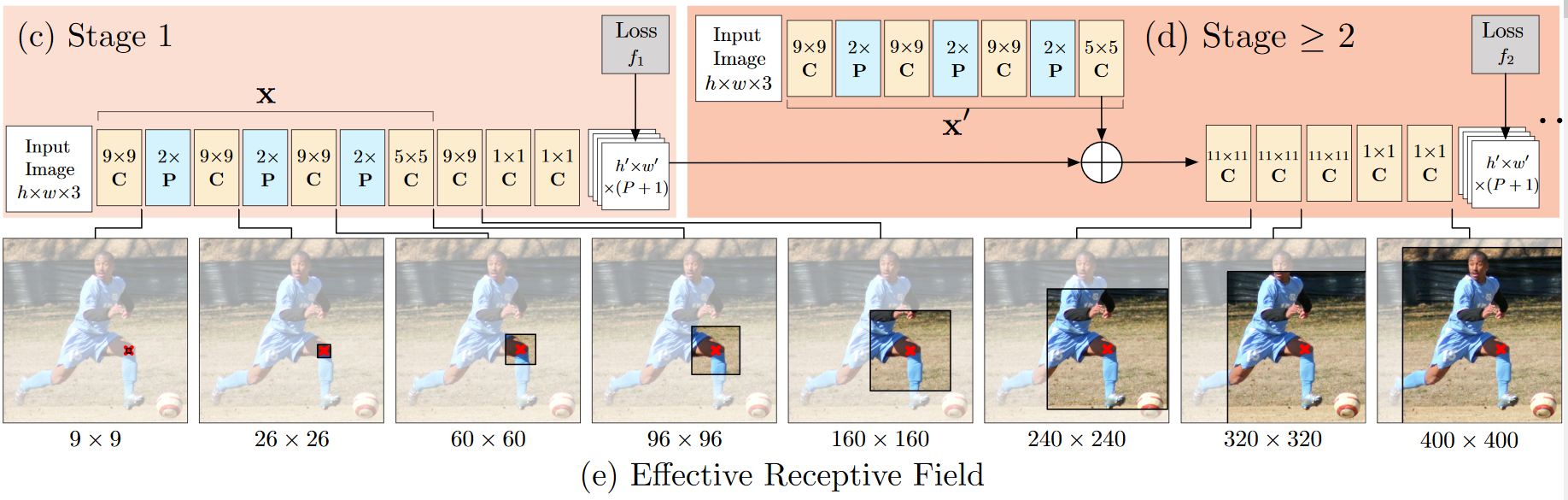}}
\caption{Convolutional Pose Machine by Wei et al.~\cite{Wei} and the ERF of neurons in different layers (Figure taken from~\cite{Wei}).}
\label{fig:PoseMachine}
\vspace*{-10pt}
\end{figure*}

\begin{figure*}[t]
\centerline{\includegraphics[width=.7\textwidth]{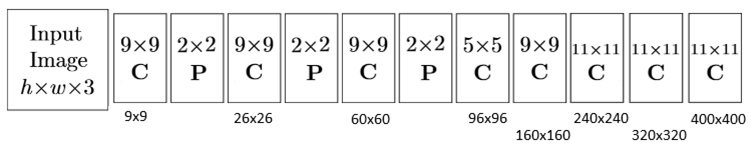}}
\caption{Sample CNN architecture with filter sizes and effective receptive fields shown, reproduced from \cite{Wei}.}
\label{fig:Architecture}
\vspace*{-10pt}
\end{figure*}

Here, we calculate the ERF of neurons for an the CNN from Wei et al.,~\cite{Wei}. In their paper, Wei et al. proposed a method for pose estimation (known as Convolutional Pose Machine). Figure~\ref{fig:PoseMachine} shows the original architecture. Here, we focus on calculating ERF for part of the network shown in Figure~\ref{fig:Architecture}, with the 1x1 filters being omitted.

\noindent \textbf{Bottom-Up approach:} The ERF of each layer is computed progressively while skipping the $1\times1$ filters as they do not have any effect on the ERF size. The process of computing ERF of the architecture in Figure \ref{fig:Architecture}, according to equation \ref{eq:eq3}, is shown below:

\begin{equation}
\begin{matrix*}[l]
R_{0} = 1 \\
R_{1} = 1 + (9 - 1)(1) = 9 \\
R_{2} = 9 + (2 - 1)(1) = 10\\
R_{3} = 10 + (9 - 1)(2) = 26\\
R_{4} = 26 + (2 - 1)(2) = 28\\
R_{5} = 28 + (9 - 1)(2*2) = 60\\
R_{6} = 60 + (9 - 1)(2*2) = 64\\
R_{7} = 64 + (5 - 1)(2*2*2) = 96\\
R_{8} = 96 + (9 - 1)(2*2*2) = 160\\
R_{9} = 160 + (11 - 1)(2*2*2) = 240\\
R_{10} = 240 + (11 - 1)(2*2*2) = 320\\
R_{11} = 320 + (11 - 1)(2*2*2) = 400
\end{matrix*}
\label{eq:BottomUp}
\end{equation}

\noindent \textbf{Top-Down approach:} Here, the ERF is calculated for each layer separately. So for a network with n layers, n passes back to the image are needed. In other words, intermediate numbers can not be reused. The process of computing ERF for the architecture in Figure \ref{fig:PoseMachine}, according to equation \ref{eq:eq8}, is shown below. \(R_{11,0}\) is the ERF of the $11th$ layer in the network. Notice that a separate computation is needed to compute \(R_{12,0}\) or \(R_{10,0}\).

\begin{equation}
\begin{matrix*}[l]
R_{11,11} = 1 \\
R_{11,10} = (1-1)(1) + 11 = 11 \\
R_{11,9} = (11-1)(1) + 11 = 21 \\
R_{11,8} = (20-1)(1) + 11 = 31 \\
R_{11,7} = (31-1)(1) + 9 = 39 \\
R_{11,6} = (39-1)(1) + 5 = 43 \\
R_{11,5} = (43-1)(2) + 2 = 86 \\
R_{11,4} = (86-1)(1) + 9 = 94 \\
R_{11,3} = (94-1)(2) + 2 = 188 \\
R_{11,2} = (188-1)(1) + 9 = 196 \\
R_{11,1} = (196-1)(2) + 2 = 392 \\
R_{11,0} = (392-1)(1) + 9 = 400
\end{matrix*}
\label{eq:TopDown}
\end{equation}

\section{Projective field size}
In this section, we discuss the calculation of the PF size of a neuron.
For the example in Section~\ref{sec1}, assuming 10 filters of size $5 \times 5$, and stride equal 5, in the first convolutional layer, the PF of each image pixel (i.e., input neuron and in each R, G, or B channels), would be $1\times1\times10$. Notice that this calculation is independent of the filter size but as we will show below depends on the stride size. Further, notice that as in calculation of the RF, there is a depth component involved as well. For simplicity, we discard the depth in what follows.

\begin{figure}[t]
\begin{center}
\includegraphics[width=0.5\linewidth]{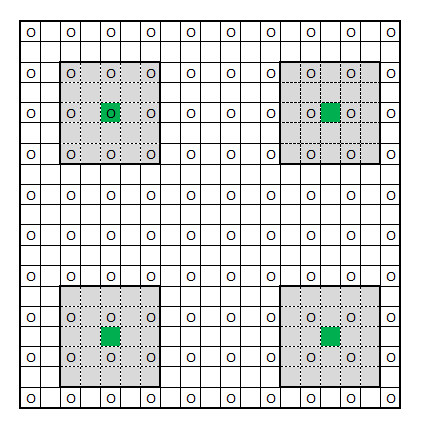}
\end{center}
   \caption{Projective field size. The gray box is the filter with size of $5 \times 5$. The small circle is the output node of the filter map with the stride of two. The PF of a neuron (green box) is calculated by counting how many filters (circles) being applied within the bounding box. Depend on the location, the PF of a neuron would be different from other.}
\label{fig:ProjectiveField}
\end{figure}

\begin{figure}[t]
\begin{center}
\includegraphics[width=0.5\linewidth]{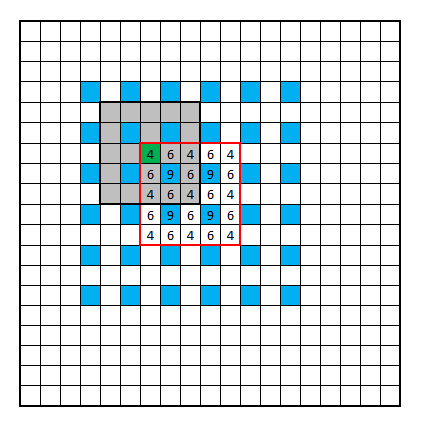}
\end{center}
   \caption{Illustration of projective field size calculation. In this image, the blue box is where the filter is being applied. To calculate the PF of a neuron (shown in green), the filter (gray area) is applied around the neuron to determine the PF. For the showed neuron, the PF is 2x2. The same action can be done with other cell. The values can be verified with more complex process showed in Figure~\ref{fig:Grid1}, and Figure~\ref{fig:Grid2}.}
\label{fig:PFCounting}
\end{figure}

\begin{figure*}[t]
\centerline{\includegraphics[width=0.95\textwidth]{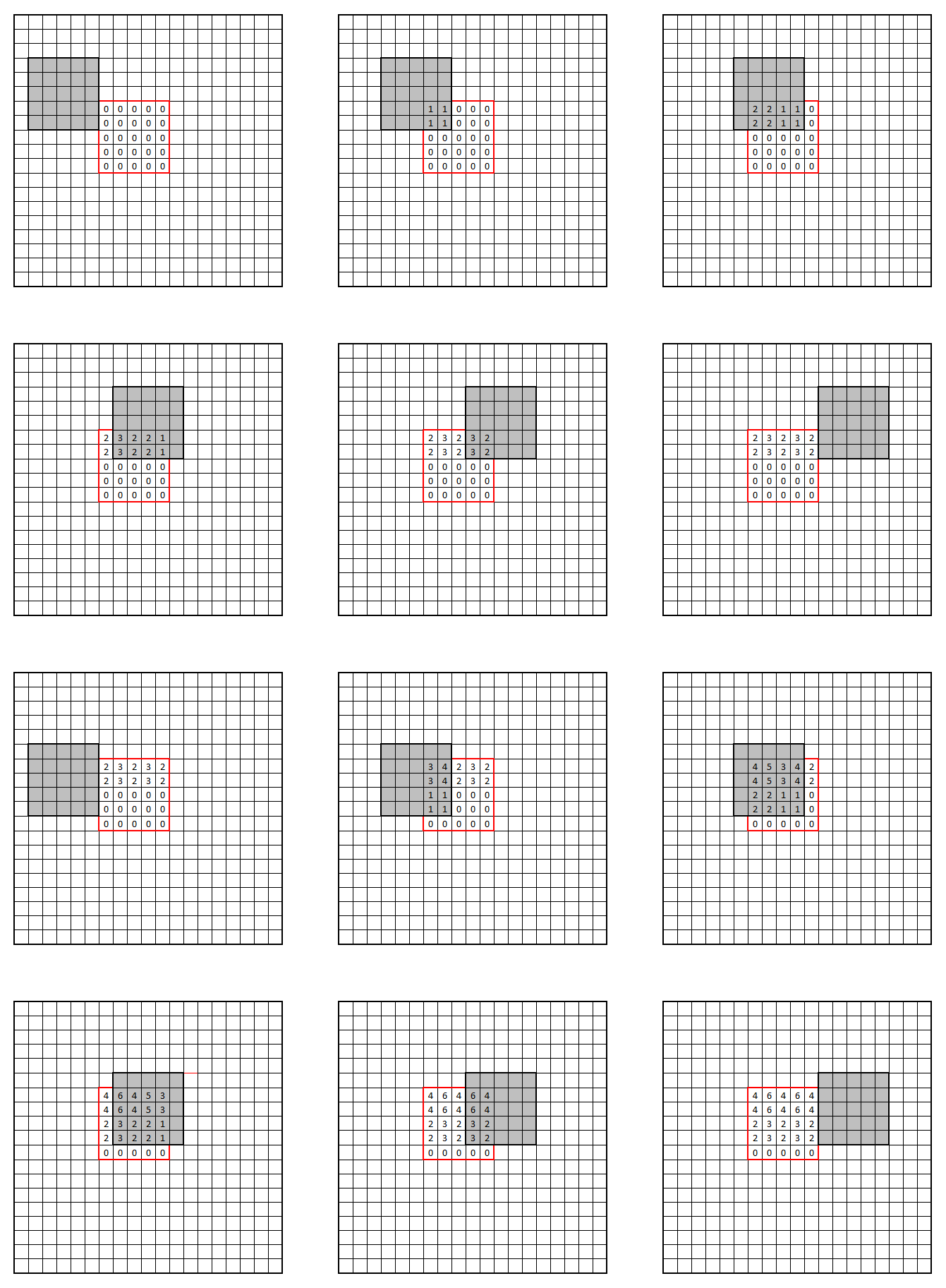}}
\caption{Sequence of sliding a filter for calculating the projective field of neurons. Here filter size is $5 \times 5$ and stride is 2.}
\label{fig:Grid1}
\vspace*{-10pt}
\end{figure*}

\begin{figure*}[t]
\centerline{\includegraphics[width=0.95\textwidth]{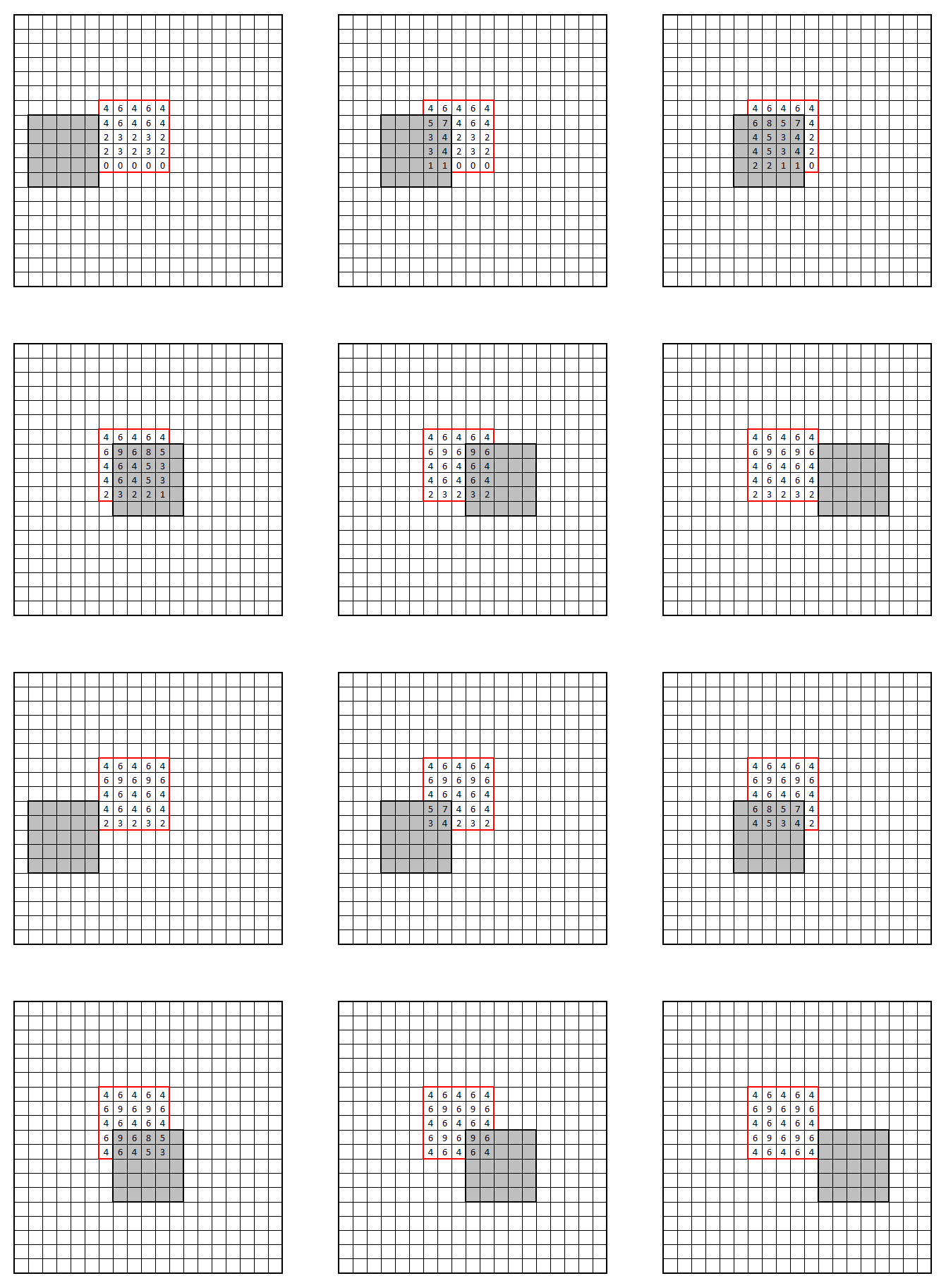}}
\caption{Sequence of sliding a filter for calculating the projective field of neurons (continued).}
\label{fig:Grid2}
\vspace*{-10pt}
\end{figure*}

The size of the projective field can be calculated by sliding the filter over an area and update the counter of each neuron when it overlaps with the filter (Figure \ref{fig:Grid1}, \ref{fig:Grid2}). However, sliding the filter is prone to error and difficult to keep track of the value in the x and y directions. A simpler method to determine the projective field was is shown in Figures \ref{fig:ProjectiveField} and \ref{fig:PFCounting}.

With a stride of 1, the immediate PF is the same size as the filter size of the next layer. For example, if the filter size is $3 \times 3$, then a neuron will influence $3 \times 3$ nodes in the output filter map. For the above example, assuming 10 filters in the first convolutional layer, the PF of each image pixel (i.e., input neuron and in each R, G, or B channel), would be $3\times3\times10$. The pixels at the corners and the edges would have slightly smaller PFs. Here, for simplicity we assume that the input image has been zero padded. 

With a stride greater than 1, then some neurons will have bigger PFs than  other neurons. For example, for a filter size of $5 \times 5$ and stride of 2, the center neuron (see figure \ref{fig:ProjectiveField}) has the PF of $3 \times 3$. The neurons on the x-axis and y-axis of the center neuron would have PFs of $3 \times 2$ or $2 \times 3$ , respectively. The neurons diagonal to the center neurons would have PF sizes of $2 \times 2$.

From the above analysis (See Figures \ref{fig:Grid1}, \ref{fig:Grid2}, and \ref{fig:PFCounting}), the projective field of a node at layer k of a CNN is bounded with a set of four pairs of values:
\begin{equation}
\begin{matrix*}[l]
P_k = \bigg \{ \floor*{\frac{f_{k+1}}{s_{k+1}}}\times\floor*{\frac{f_{k+1}}{s_{k+1}}}, 
\floor*{\frac{f_{k+1}}{s_{k+1}}}\times\ceil*{\frac{f_{k+1}}{s_{k+1}}}, \\\\
\ceil*{\frac{f_{k+1}}{s_{k+1}}}\times\floor*{\frac{f_{k+1}}{s_{k+1}}}, 
\ceil*{\frac{f_{k+1}}{s_{k+1}}}\times\ceil*{\frac{f_{k+1}}{s_{k+1}}} \bigg \}
\end{matrix*}
\label{eq:ProjectiveField}
\end{equation}
where $f_{k+1}$ and $s_{k+1}$ are the filter size and stride in the next layer. According to the equation \ref{eq:ProjectiveField}, if the remainder of the fraction is zero, then all the nodes have equal projective field sizes. Otherwise, depending on the location of the nodes, projective field sizes would be different. In other words, when the fraction does not yield an integer value, there is disparity in the influence of the nodes in the next layer. This is perhaps why researchers tend not to use strides greater than 1 in convolution layers (or use strides equal the filter size in pooling layers). Nonetheless, it is unclear whether such disparity can cause any practical problems. 

Deconv nets are inverted versions of CNNs. Therefore, their projective field can be calculated using the ERF formulas. Similarly, their ERF is the same as the projective field in CNNs.

\section{Discussion}
Here, we discussed how receptive, effective receptive and projective fields of neurons in CNNs are calculated. Understanding these quantities has important implications in deep learning research. First, it helps setting the parameters such as filter size, number of filters, and stride more effectively. Second, it allows analyzing how objects are represented by CNNs and investigate whether some image information is lost along the CNN hierarchy.

\vspace*{-5pt}
\section{Acknowledgment}
We wish to thank all participants in the Advanced Computer Vision course at UCF who contributed to discussions.

\vspace*{-5pt}
{\small
\bibliographystyle{ieee}
\bibliography{references}

\begin{thebibliography}{1}\itemsep=-1pt

\bibitem{hubel1962receptive}
D.~H. Hubel and T.~N. Wiesel.
\newblock Receptive fields, binocular interaction and functional architecture
  in the cat's visual cortex.
\newblock {\em The Journal of physiology}, 160(1):106--154, 1962.

\bibitem{lecun1998gradient}
Y.~LeCun, L.~Bottou, Y.~Bengio, and P.~Haffner.
\newblock Gradient-based learning applied to document recognition.
\newblock {\em Proceedings of the IEEE}, 86(11):2278--2324, 1998.

\bibitem{lehky1988network}
S.~R. Lehky and T.~Sejnowski.
\newblock Network model of shape-from-shading: neural function arises from both
  receptive and projective fields.
\newblock {\em Nature}, 333(6172):452--454, 1988.

\bibitem{Wei}
S.-E. Wei, V.~Ramakrishna, T.~Kanade, and Y.~Sheikh.
\newblock Convolutional pose machines.
\newblock In {\em 2016 IEEE Conference on Computer Vision and Pattern
  Recognition (CVPR)}, pages 4724--4732, June 2016.

\end{thebibliography}
}
\end{document}